\documentclass[10pt,twocolumn,letterpaper]{article}

\usepackage[pagenumbers]{cvpr} 
\usepackage[accsupp]{axessibility}  

\definecolor{cvprblue}{rgb}{0.21,0.49,0.74}
\usepackage[pagebackref,breaklinks,colorlinks,allcolors=cvprblue]{hyperref}

\usepackage{enumitem}
\usepackage{bm}
\usepackage{booktabs}
\usepackage{xcolor}
\usepackage{threeparttable}
\renewcommand{\paragraph}[1]{\noindent\textbf{#1}\ \ }
\usepackage{xcolor}
\usepackage{colortbl}
\definecolor{rank1}{RGB}{226, 164, 145} 
\definecolor{rank2}{RGB}{235, 197, 185} 
\definecolor{rank3}{RGB}{244, 227, 222} 
\usepackage{multirow}

\def\paperID{****} 
\def\confName{CVPR}
\def\confYear{2025}

\title{
Towards Enhanced Image Inpainting: \\
Mitigating Unwanted Object Insertion and Preserving Color Consistency
}

\author{
Yikai Wang$^{1,2}$\footnotemark[1], Chenjie Cao$^{1,3,4*}$, Junqiu Yu$^{1*}$, Ke Fan$^{1}$, Xiangyang Xue$^{1}$, Yanwei Fu$^{1}$ \\
$^{1}$Fudan University\quad $^{2}$Nanyang Technological University\quad$^{3}$Alibaba DAMO Academy\quad$^{4}$Hupan Lab\\
{\tt\small yi-kai.wang@outlook.com, yanweifu@fudan.edu.cn}\\
\small{Project page (include code, model, and dataset): \url{https://yikai-wang.github.io/asuka}}
}

\begin{document}
\maketitle
\renewcommand{\thefootnote}{\fnsymbol{footnote}}
\footnotetext{
First three authors contribute equally. 
Most parts of this work was done when Yikai was at Fudan.
Yanwei Fu is the corresponding author.
} 

\begin{abstract}
Recent advances in image inpainting increasingly use generative models to handle large irregular masks. 
However, these models can create unrealistic inpainted images due to two main issues:
(1) \textbf{Unwanted object insertion}: 
Even with unmasked areas as context, generative models may still generate arbitrary objects in the masked region that don’t align with the rest of the image.
(2) \textbf{Color inconsistency}: Inpainted regions often have color shifts that causes a smeared appearance, reducing image quality.
Retraining the generative model could help solve these issues, but it’s costly since state-of-the-art latent-based diffusion and rectified flow models require a three-stage training process: training a VAE, training a generative U-Net or transformer, and fine-tuning for inpainting.
Instead, this paper proposes a post-processing approach, dubbed as ASUKA (Aligned Stable inpainting with UnKnown Areas prior), to improve inpainting models.
To address unwanted object insertion, we leverage a Masked Auto-Encoder (MAE) for reconstruction-based priors. 
This mitigates object hallucination while maintaining the model's generation capabilities.
To address color inconsistency, we propose a specialized VAE decoder that treats latent-to-image decoding as a local harmonization task, significantly reducing color shifts for color-consistent inpainting.
We validate ASUKA on SD 1.5 and FLUX inpainting variants with Places2 and MISATO, our proposed diverse collection of datasets. 
Results show that ASUKA mitigates object hallucination and improves color consistency over standard diffusion and rectified flow models and other inpainting methods.
\end{abstract}

\section{Introduction}
\begin{figure}
\centering
\includegraphics[width=\linewidth]{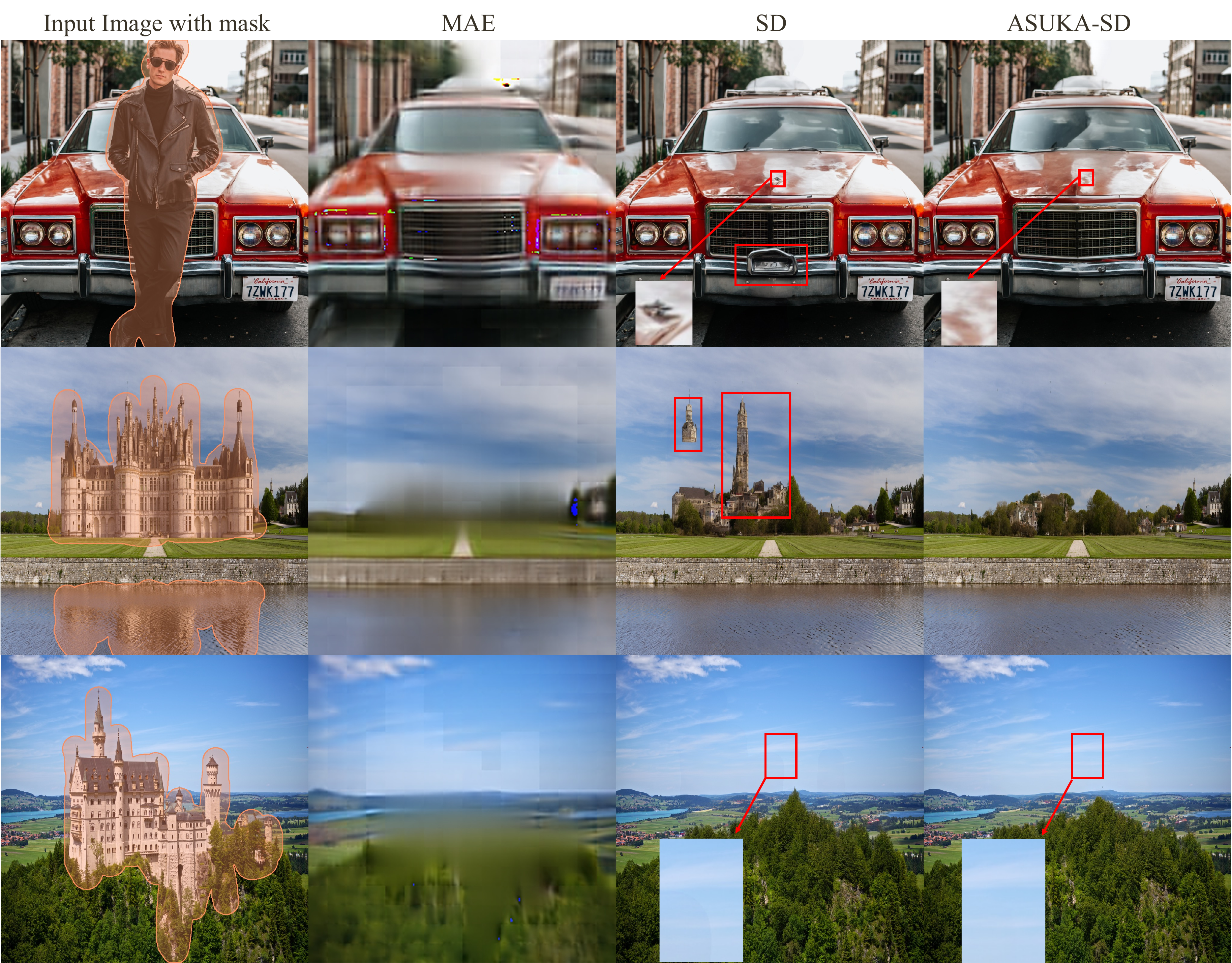}
\vspace{-0.5cm}
\caption{
Image inpainting on $1024^2$ images.
ASUKA solves two issues existed in current diffusion and rectified flow inpainting models:
(1) Unwanted object insertion, where randomly elements that are not aligned with the unmasked region are generated;
(2) Color-inconsistency: the color shift of the generated masked region, causing smear-like traces.
ASUKA proposes a post-training procedure for these models, significantly mitigates object hallucination
and improves color consistency of inpainted results.
}
\label{fig:object_removal}
\vspace{-0.5cm}
\end{figure}
Image inpainting~\cite{bertalmio2000image} fills masked areas of images while maintaining consistency with the unmasked regions. 
Traditional inpainting algorithms~\cite{bertalmio2000image,Criminisi2003ObjectRB,hays2007scene,Levin2003LearningHT,Roth2005FieldsOE} often result in blurred synthesis when reconstructing masked regions~\cite{pathak2016context}.
The Generative Adversarial Networks (GANs) based models could fill complex mask structures, achieving impressive inpainting results~\cite{nazeri2019edgeconnect,liao2020guidance,cao2021learning,suvorov2022resolution,wan2021highfidelity,zhao2021large,goodfellow2014generative,ho2020denoising,esser2021taming}. 
However, they still struggle with general challenging inpainting cases, particularly in filling large holes.
Recently, more powerful generative model like Stable Diffusion~\cite{Rombach_2022_CVPR} and FLUX~\cite{flux}, with their large model capacity and extensive training dataset, act as versatile tools for image inpainting. 
These models mainly follow the latent generation pipeline, first encode the image into a small latent space, then train the  inpainting model in this latent space. 

However, these latent-based generative inpainting models still suffer from some issues, causing the inpainted image lacking fidelity. In particular:

\noindent(1) The \textit{unwanted object insertion} problem, where the model generates random, unreasonable elements to fill masked regions, as depicted in first to second rows in Fig.~\ref{fig:object_removal}.
This issue comes from the random masking strategy used to train generative models.
This strategy introduces training cases where foreground objects are completely masked but the models are forced to fill masked regions with foreground objects. 
Consequently, these models will hallucinate unreasonable objects devoid of contextual information. 
Adjusting prompts may reduce this risk, but the best prompt is image-dependent, making it infeasible for practical applications. 

\noindent(2) Furthermore, the inpainted results of latent inpainting models suffer from \textit{color inconsistency} problem.
This problem, less explored in academia but critical for real-world applications, results in color discrepancies between inpainted and unmasked regions, including mismatches of brightness, saturation, luminance and hue, and exhibits smear-like traces in the image, as shown in the second to third rows in Fig.~\ref{fig:object_removal}.
Essentially, this color-inconsistency comes from the misalignment between the pixel distributions of filled results and original images due to imperfect latent generative model and VAE, as illustrated in Fig.~\ref{fig:color-shift}. 
Notably, this issue is not a big problem for generation, given that the whole image is generated, and the color shift is consistent across pixels.
However, this issue is important for inpainting tasks, as we have ground-truth pixels for unmasked regions.
When we replace the unmasked regions of the generated image with the ground-truth pixels, the color inconsistency largely influence the fidelity of the image.
This issue may be solved by training a better VAE and explicitly enforce the color consistency.
However, training or fine-tuning the VAE encoder introduces the subsequent fine-tuning of the latent generative models to match the new latent space, which is costly.
In this paper, we propose to freeze the VAE encoder and the latent generative models, while fine-tuning the VAE decoder to improve color consistency.
Specifically, we reformulate the decoding from latent to image as a local harmonization task, explicitly reduce the color inconsistency.

Formally, to mitigate object hallucination and  enhance the color-consistency of  image inpainting models, we present the Aligned Stable inpainting with UnKnown Areas prior (ASUKA) framework.
ASUKA enhances the latent inpainting models with regression-based reconstruction and distribution-aligned generation. 
This results in improved image inpainting models that avoid generating unreasonable elements in the masked region and reduces mask-unmask color inconsistency. 
The stable diffusion models~\cite{Rombach_2022_CVPR} and the rectified flow models~\cite{flux} adopt a VAE to compress image into latent and perform inpainting in the latent space.
We manipulate their generation and decoding processes to reduce object hallucination
and improve color consistency. 

We propose using the Masked Auto-Encoder (MAE)\cite{he2022masked} as a prior to guide and stabilize the generation process. 
As shown in Fig.~\ref{fig:object_removal}, MAE yields stable yet blurred results, while generative models may produce implausible content despite their impressive generation capacity. 
By aligning MAE prior with latent generative models, we reduce object hallucination without damaging performance. 

We redesign the VAE decoder to address color inconsistencies between masked and unmasked regions by acting as a local harmonization model conditioned on unmasked image pixels.
\emph{Our decoder can be used as a plug-and-play module to improve general inpainting models, such as text-guided inpainting}.

These steps collectively enable ASUKA to achieve less object hallucination and more color-consistent inpainting results.
We adopt ASUKA on two typical inpaitning models,
Stable Diffusion v1.5~\cite{Rombach_2022_CVPR} and FLUX~\cite{flux},
to validate the generalization ability of ASUKA on different generation architectures.
To evaluate the effectiveness of inpainting algorithms across various scenarios and mask shapes, in addition to the benchmark dataset Places 2~\cite{zhou2017places}, we further utilize an evaluation dataset named MISATO, which selects representative testing images from \underline{M}atterport3D~\cite{Matterport3D}, 
Fl\underline{i}ckr-Land\underline{s}cape~\cite{lin2021infinity},
Meg\underline{a}Dep\underline{t}h~\cite{MDLi18}, and C\underline{O}CO 2014~\cite{lin2014microsoft}.
This dataset covers four distinct domains—landscape, indoor, building, and background—making it diverse to serve as a benchmark for evaluation. 
Experiments on MISATO and Places 2 with large irregular masks validate the efficacy of ASUKA.

\paragraph{Contributions} 
ASUKA enhances image inpainting with color-consistency and mitigate object hallucination while leveraging the generation capacity of the frozen inpainting model. It achieves this through two main components: 
\textit{(1) Context-Stable Alignment}: ASUKA aligns the stable MAE prior with generative models to provide a context-stable estimation of masked regions, replacing the text-condition with MAE prior. 
\textit{(2) Color-Consistent Alignment}: ASUKA re-formulates the decoding from latent to image as a local harmonization task, trains an inpainting-specialized decoder to align masked and unmasked regions during decoding and thus mitigates color inconsistencies. 

\section{Related Works}

\paragraph{Image inpainting} is the task of filling missing image regions with consistent pixels. Traditional methods using patch matching~\cite{criminisi2004region,barnes2009patchmatch,zhang2018robust} or differential equations~\cite{bertalmio2000image,chan2001nontexture,bertalmio2003simultaneous} focus on low-level features and often struggle with large gaps. 
GAN~\cite{goodfellow2014generative}-based inpainting~\cite{pathak2016context,yu2019free,zhao2021large,li2022mat,cao2022learning} introduces adaptive convolutions~\cite{liu2018image,yu2019free,zeng2022aggregated}, attention~\cite{yu2018generative,yi2020contextual,zeng2020high,ko2023continuously}, and frequency-based learning for high-resolution results~\cite{suvorov2022resolution,xu2023image,chu2023rethinking}. 
Methods like Co-Mod~\cite{zhao2021large} address the challenging ill-posed inpainting issue~\cite{li2022mat,zheng2022image} and improve realism but may produce unstable outputs or unwanted artifacts due to random latent variables. 
Techniques with higher reconstruction penalties~\cite{nazeri2019edgeconnect,suvorov2022resolution,cao2022learning} offer more stability but can appear blurry on larger missing areas. 
Recent diffusion models~\cite{saharia2022palette,Rombach_2022_CVPR,arkhipkin2023kandinsky} and rectified flow models~\cite{esser2024scaling, flux} achieve impressive results yet share GANs' limitation of learning distributions over exact pixel alignment, which leads to unwanted object insertion.

\paragraph{Adapting latent generative models} 
Latent diffusion models (LDMs)~\cite{Rombach_2022_CVPR} and rectified flow models~\cite{esser2024scaling, flux} are popular due to their ability to encode image semantics at lower resolutions by combining a VAE to learn a latent space and a generative model within this space. 
Various methods have been developed to introduce new conditions to these models, such as image-inversion for text-guided image translation~\cite{meng2021sdedit}, textual-inversion for personalization~\cite{gal2022image}, LoRA fine-tuning~\cite{hu2021lora}, and controlnet~\cite{zhang2023adding} to add diverse conditions. 
Our goal is to mitgate object hallucination while preserving generation quality, so we avoid fine-tuning the generative backbone. 
For inpainting, we remove the text condition and instead guide the generation using a Masked Auto-Encoder~\cite{he2022masked} prior for masked regions.

\paragraph{Information loss in latent inpainting models}
Although claimed only eliminates imperceptible details, the VAE used by diffusion and rectified flow models causes distortion in the reconstruction of images.
In addition, the gap between generated latent and real latent also causes the color inconsistency.
See Fig.~\ref{fig:color-shift} for illustrative examples.
OpenAI~\cite{openai-decoder} proposes a larger decoder to improve the decoding quality of SD's latent.
Luo \emph{et. al.}~\cite{luo2023image} propose a frequency-augmented decoder to address the super-resolution case.
Zhu \emph{et. al.}~\cite{zhu2023designing} propose to preserve unmasked regions during decoding.
In this paper, we ensure the low-frequency color color-consistency in the decoding process.

\paragraph{Masked Image-Modeling~\cite{bao2021beit}} (MIM) is an active research area in self-supervised learning. 
Typical MIM methods~\cite{bao2021beit, he2022masked,xie2022simmim,chen2023context} split images into visible and masked patches, learning to estimate masked patches from visible patches. 
Training targets for visible patches encompass  pixel values~\cite{he2022masked}, HOG features~\cite{wei2022masked}, and high-level semantic features~\cite{wei2022mvp}.
While  the primary objective of MIM is representation learning, its potential effectiveness in image generation is also noteworthy.
Cao \emph{et. al.}~\cite{cao2022learning} adopts MAE features and attention scores to assist the convolutional inpainting model better in handling long-distance dependencies. In contrast, this paper uses MAE prior to enhance the context-stability of diffusion and rectified flow models.

\paragraph{Image harmonization} aims to blend a foreground object with a background image while keeping the final result realistic and visually consistent~\cite{tsai2017deep}. 
This task is often treated as an image translation problem~\cite{zhu2015learning, cong2020dovenet, guo2021intrinsic, cong2022high,guo2022transformer, niu2023deep, wang2023semi, liu2023lemart,meng2024high, ren2024relightful}.
Similarly, our work addresses color inconsistency issues in latent generative models. 
However, unlike standard image harmonization, where inconsistencies arise from combining images from different sources and thus different real image distributions, color inconsistencies in latent generative models stem from imperfections in the VAE and the generative model itself.

\paragraph{Object insertion and removal}
are two opposite tasks in image inpainting. 
Object insertion focuses on adding foreground objects to the image using various methods, such as shape-guided masks~\cite{zeng2022shape,xie2023smartbrush}, text prompts~\cite{wang2023imagen,xie2023smartbrush,canberk2024erasedraw,chiu2024brush2prompt}, learnable prompts~\cite{wang2024repositioning,zhuang2024task,chiu2024brush2prompt}, extra network modules~\cite{chen2024improving, ju2024brushnet}, or reference images of objects~\cite{saini2024invi}, etc. 
Some studies also explore completing partial objects using reference images~\cite{cao2024leftrefill} or learnable prompts~\cite{wang2024repositioning}.
Object removal, on the other hand, aims to erase unwanted objects from an image. 
Common approaches include attention reweighting~\cite{li2024magiceraser} and learnable prompts~\cite{wang2024repositioning,zhuang2024task}. 
These techniques can help create new datasets~\cite{de2024placing}.
On the other hand, creating new datasets can also benefit these tasks~\cite{winter2024objectdrop}.
While most research focuses on designing better inpainting models, our work takes a different approach. We analyze a fundamental problem with latent generative models: they often introduce unwanted objects in the inpainting area. 
We also propose solutions to address this issue.

\section{Methodology}
\paragraph{Problem setup}
Inpainting takes as inputs a masked image to complete with a mask to indicate the missing region. 
The target of inpainting is to fill the missing region based on the information of unmasked regions to generate high-fidelity images.
In this paper, we focus on the standard inpainting task without utilizing other conditions.
We focus on the general issues of inpainting models, 
(1) \textbf{unwanted object insertion}: unstable and uncontrollable hallucinations, yielding random elements generated in the masked region;
(2) \textbf{color-inconsistency}: mask-unmask color inconsistency issue, yielding smear-like traces in the masked region.

\begin{figure}
\centering
\includegraphics[width=1\linewidth]{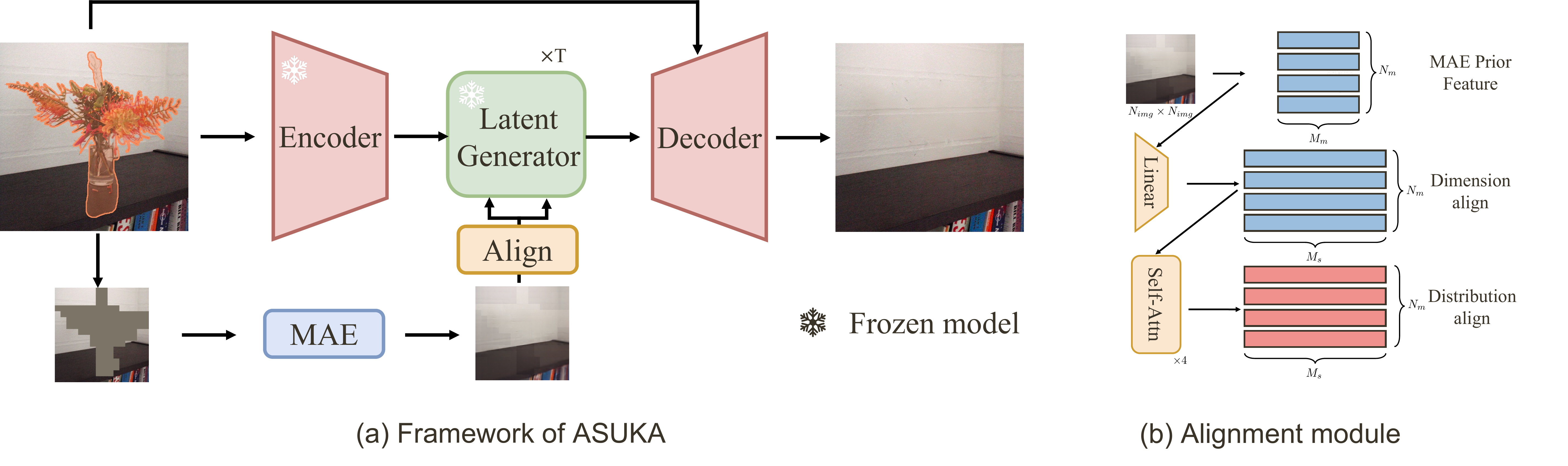}
\vspace{-0.5cm}
\caption{
ASUKA tackles the unwanted object insertion issue by adopting the MAE to provide a stable prior for frozen latent generative models to maintain the generation capacity while mitigating object hallucination.
For the color-inconsistency issue, ASUKA utilizes an inpainting-specialized decoder to achieve mask-unmask color consistency when decoding latent.
}
\label{fig:asuka}
\vspace{-0.5cm}
\end{figure}

We evaluate our proposed solution on two inpainting models: the Stable Diffusion v1.5 inpainting model (SD)~\cite{Rombach_2022_CVPR} and the Control-Net fine-tuned FLUX inpainting model (FLUX)~\cite{flux-inpainting}. 
We provide a brief introduction of these models in the appendix.
We will demonstrate that our ASUKA effectively improves unwanted object mitigation and color consistency of these  models.

\paragraph{Overview}
The framework of the proposed Aligned Stable inpainting with Unknown Areas prior (ASUKA) is illustrated in Fig.~\ref{fig:asuka}(a).
ASUKA adopts the pre-trained latent inpainting models.
Our target is to mitigate object hallucination and provide more color-consistent inpainting results while fully exploiting the generation capacity of frozen models.
ASUKA includes 
\textbf{(1)} a \textit{context-stable alignment} to align stable Masked Auto-Encoder (MAE) prior for masked region with generative models and 
\textbf{(2)} a \textit{color-consistent alignment} to align ground-truth unmasked region with generated masked region during decoding.
To this end, we freeze the latent generative models, while replacing the text-condition part with our proposed  MAE prior to mitigate object hallucination.
To align the MAE prior to generative models, we introduce an alignment module, trained via the training objective of generative models. 
Additionally, to align masked and unmasked regions during decoding and resolve the information loss issue from VAE decoder and generative model which causes mask-unmask color inconsistency, we train an inpainting-specialized decoder to decode the latent back to the image space for seamless color-consistency.
Combined together, ASUKA achieves less object hallucination and more color-consistent inpainting.

\subsection{Mitigate Object Hallucination via Stable Prior}
\subsubsection{Masked Auto-Encoder Prior}
\paragraph{Context-stable prior} 
While recent generative models rely on random noise to provide more diverse generation results, it leads to the generation of random objects unexpectedly. 
Some inpainting models also utilize the reconstruction loss to reconstruct the masked region, but they also incorporate other types of losses like perceptual-loss~\cite{suvorov2022resolution} which implicitly reduces the stability.
In contrast, MAE is known to provide a context-stable estimation of masked regions based purely on the unmasked regions.
In this paper, we utilize MAE to produce the stable prior such that 
\emph{the improvement of inpainted result can be explicitly attributed to the improvement of mitigating object hallucination}.

\begin{figure}
\centering
\includegraphics[width=0.8\linewidth]{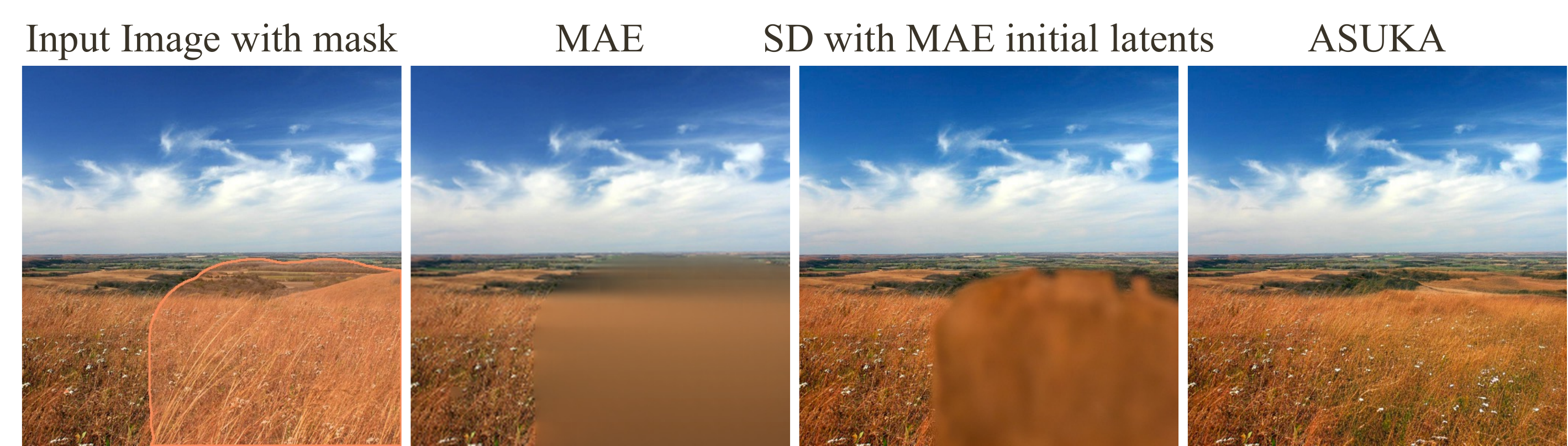}
\caption{
\label{fig:mae-to-image}
Use MAE prior for image-to-image translation (start from 80\% noise rate) via SD achieves poor inpainting results.
}
\vspace{-0.5cm}
\end{figure}

\paragraph{MAE as context-stable prior} 
As MAE is trained on the L2 reconstruction loss, we can regard the estimation of MAE as a mean estimation, which can be utilized to provide a context-stable prior for generative models to not generate new concepts.
However, MAE itself results in average and blur generations and cannot reconstruct detailed textures of the masked region, and works poorly if we use MAE prior as the initial values for the inpainting models to inpaint in image-to-image style, as in Fig.~\ref{fig:mae-to-image}.
To this end, we adopt the MAE to provide prior to stabilizing diffusion models.

\label{sec:mask-strategy}
\paragraph{Train MAE}
The original MAE is trained to estimate random masks uniformly distributed in the image, while inpainting task usually contains large continuous masks.
Inspired by Cao \emph{et. al.}~\cite{cao2022learning}, we fine-tune the MAE to inpainting masks.
To adapt MAE for more practical inpainting scenarios,
we construct a systematic masking strategy.
The mask basis contains: object-shape mask, irregular mask, and regular mask.
We collect object-shape masks from COCO~\cite{lin2014microsoft} object segments.
We use irregular masks from previous studies, including Co-Mod mask~\cite{zhao2021large} and LaMa~\cite{suvorov2022resolution} mask.
The regular masks contains rectangle and complement rectangle mask.
To ensure generalization and coverage, for each mask we generate from mask basis with the probability of 50\% object-shape, 40\% irregular, and 10\% regular.
For object-shape mask basis, we combine it with irregular mask with the chance of 50\%.
This construction of mask style estimates the masks occurs in inpainting tasks, especially for the object removal and user-specified irregular masks.
We control the mask ratio in the range of $[0.1, 0.75]$ to follow the training scenario of MAE.
For masks smaller than the ratio of 75\%, we enlarge the mask ratio to 75\% with randomly selected mask regions.
This benefits ASUKA to tackle the large hole inpainting task.

\subsubsection{Align MAE Prior with Generator}
\paragraph{Replace text-condition with MAE prior}
Generative inpainting models are not trained on MAE priors.
As we do not assume a text condition for inpainting task,
we propose to replace the text-condition of generative models with our proposed MAE prior condition.
However, as we do not fine-tune the generative models, they cannot directly align well with the MAE prior.
Hence, we introduce the alignment module to align MAE with generative models in both dimension and distribution perspective, as shown in Fig.~\ref{fig:asuka}(b).

\paragraph{Dimension alignment}
Particularly, 
the MAE prior $F_{\mathrm{MAE}}$ is of size $N_{m}\times M_{m}$, where $N_{m}$ is the sequence length and $M_{m}$ is the feature dimension.
To align it with the diffusion or flow condition of size $N_{s}\times M_{s}$,
we adopt a linear layer to map the feature dimension from $M_{m}$ to $M_{s}$ and set $N_s=N_m$ to preserve the local MAE prior.

\paragraph{Distribution alignment}
After aligning the dimension, we use self-attention blocks to learn to better guiding generative models, leading to the condition $C_{\mathrm{MAE}}$.
We train our alignment module using the standard generation objective with the same masking 
strategy used to train the MAE,
 keeping other modules frozen.

\paragraph{Handle misalignment}
When training the alignment module with the set (input image, MAE prior, inpaint result), misalignment may arise. 
For example, if an object is completely masked, the MAE will predict the masked area with background, whereas the generative models are trained to recreate the object. 
This difference can lead the alignment module to mistakenly disregard the MAE prior. 
To address this, we improve the generative models' adherence to the MAE prior by substituting the MAE predicted prior with the MAE reconstructed prior at a probability of $p$.
The MAE reconstructed prior involves using MAE to recreate the image without masking any area, ensuring MAE has access to all information needed for reconstruction. 
This approach helps train the alignment module to better guidance.

\subsection{Enhancing Color-Consistency in Decoding}
\subsubsection{Color-Inconsistency}

\begin{figure}
\centering
\includegraphics[width=0.8\linewidth]{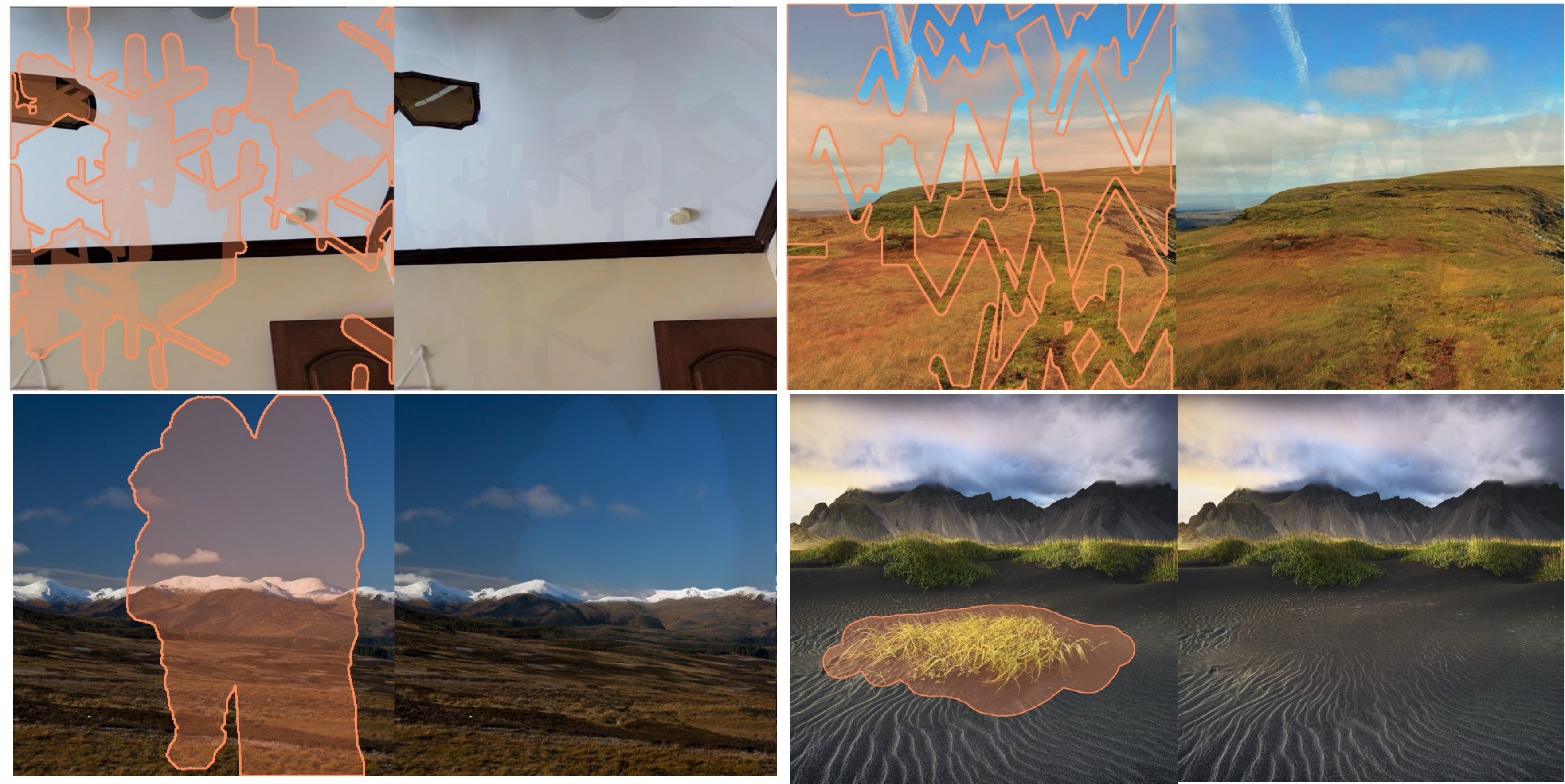}
\caption{
The color shift exists in all kinds of scenarios in inpainted images, including indoor and outdoor scenes, random or continuous masks, and may cause darker or lighter color shift.
}
\vspace{-0.2cm}
\label{fig:color-shift}
\end{figure}

\paragraph{Color-inconsistency is a general problem}
The color-inconsistency between masked and unmasked regions is a general problem in generative inpainting models.
This inconsistency comes when the generative masked region suffers from a color shift compared with the unmasked region.
As in Fig.~\ref{fig:color-shift}, the color shift happens in all kinds of scenarios, including indoor and outdoor scenes, random or
continuous masks, and may cause darker or lighter color shift.
This shift comes from the imperfect VAE and latent generator.

\begin{figure}
\centering
\includegraphics[width=\linewidth]{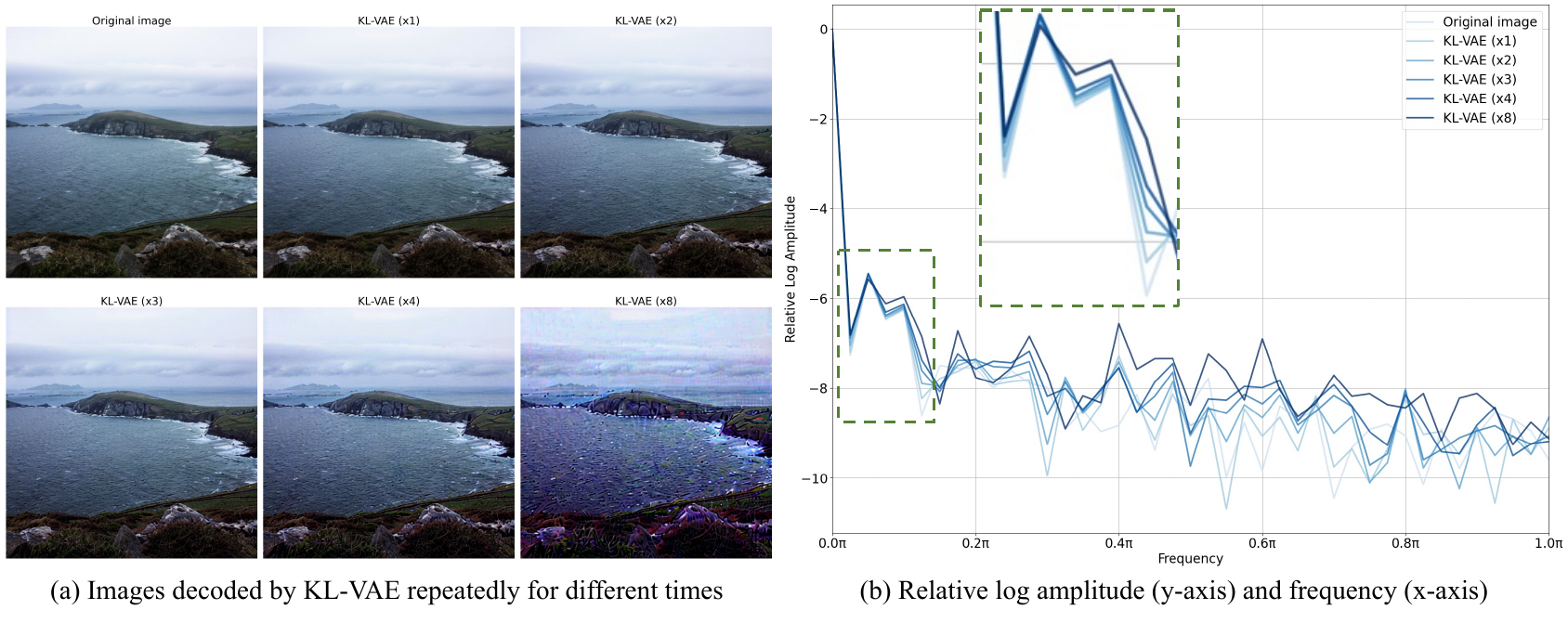}
\vspace{-0.5cm}
\caption{
(a) The color of the reconstruted image is shifted, where larger
shift is observed during repeated reconstruction. (b) VAE suffers from non-ignorable shifts in low-frequency fields.
}
\label{fig:info_loss}
\vspace{-0.2cm}
\end{figure}

\paragraph{Information loss of VAE}
Popular latent diffusion and rectified flow models perform all the generative processes in the latent space and subsequently decodes these latent codes back to image space using VAE.
Despite the decoder being trained to reconstruct the image, it encounters challenges associated with information loss.
Particularly in tasks like inpainting, we have ground-truth values for the unmasked region.
Though Rombach \emph{et. al.}~\cite{Rombach_2022_CVPR} claimed that the diffusion model should prioritize the informative semantic compression, while the VAE is used to tackle perceptual compression with high-frequency details, we argue that \textit{low-frequency semantic loss in VAE could not be neglected}, as verified in 
 Fig.~\ref{fig:info_loss} (b). 
The VAE will not only noticeably degrades high-frequency details but also shifts in colors.
This shift can be verified by repeated reconstruction with VAE, as shown in  Fig.~\ref{fig:info_loss} (a) where larger shift is observed during repeated reconstruction.
As human is sensitive to low-frequency information changes in the image, even subtle color shifts can induce significant inconsistencies. 
This issue is more severe in irregular or large mask cases.

\begin{figure}
\centering
\includegraphics[width=0.8\linewidth]{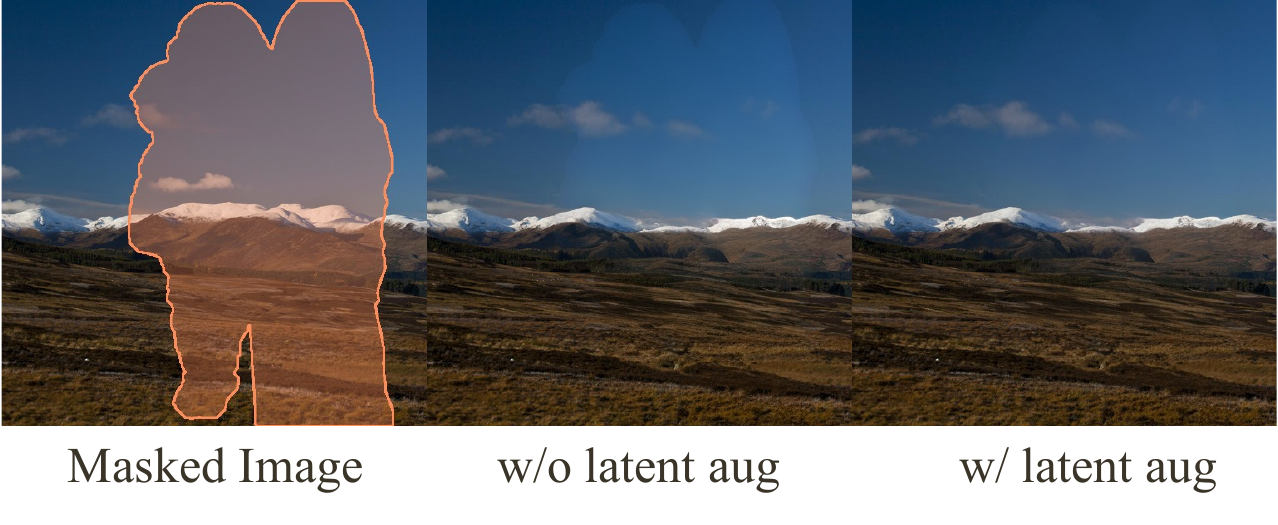}
\vspace{-0.3cm}
\caption{
 Inpainting w/ v.s. w/o latent augmentation.
The latent augmentation handles the gap between generated and real latent.
\label{fig:latent-aug-visualization}
}
\vspace{-0.2cm}
\end{figure}

\paragraph{Gap between real and generated latents}
Apart from the information loss of VAE in reconstruction, there is another gap between the generated and real latents.
This gap also causes color inconsistency even if we alleviate the VAE reconstruction loss, as in Fig.~\ref{fig:latent-aug-visualization}.
We need to solve both the loss of VAE and the latent generator for better color-consistency.

\begin{figure}
\centering
\includegraphics[width=0.8\linewidth]{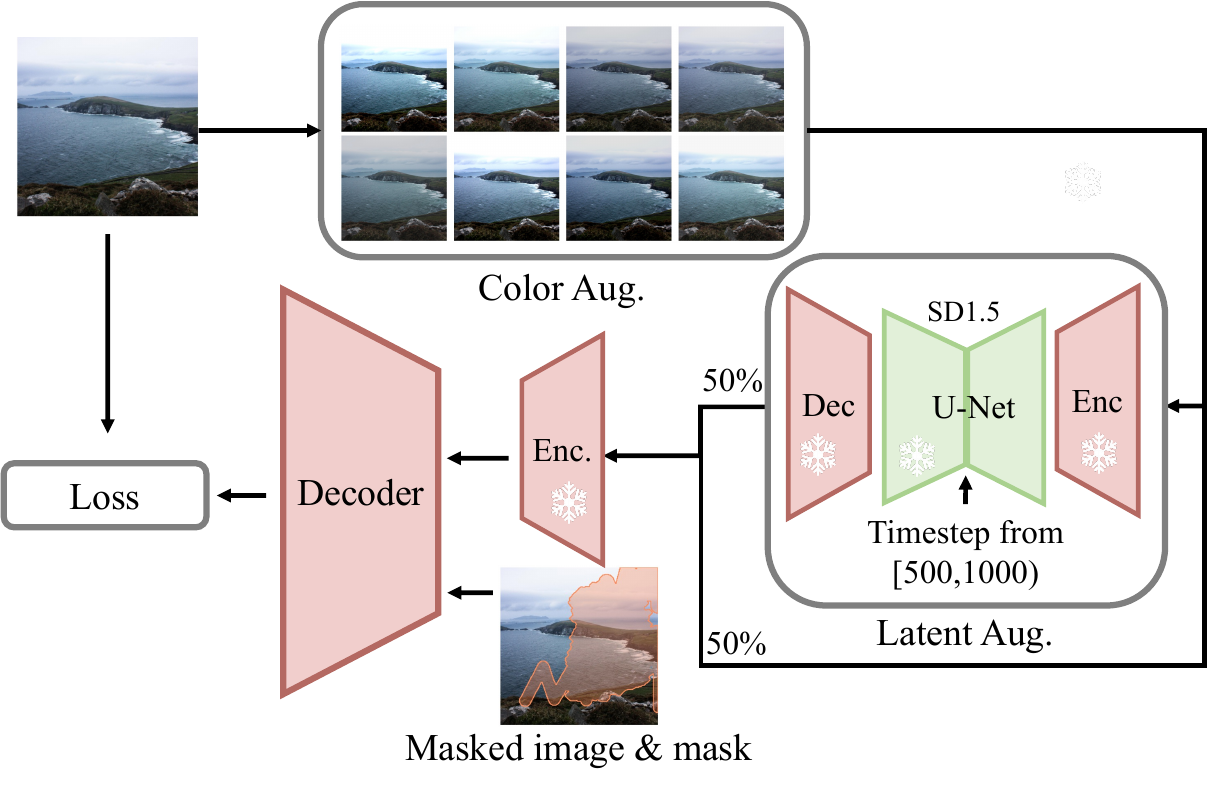}
\vspace{-0.3cm}
\caption{
Decoder trained by local harmonization task, enhancing mask-unmask consistency by reconstructing original image guided by the unmasked region from augments in color and latent spaces.\label{fig:aug}}
\vspace{-0.2cm}
\end{figure}

\begin{figure}
\centering
\includegraphics[width=\linewidth]{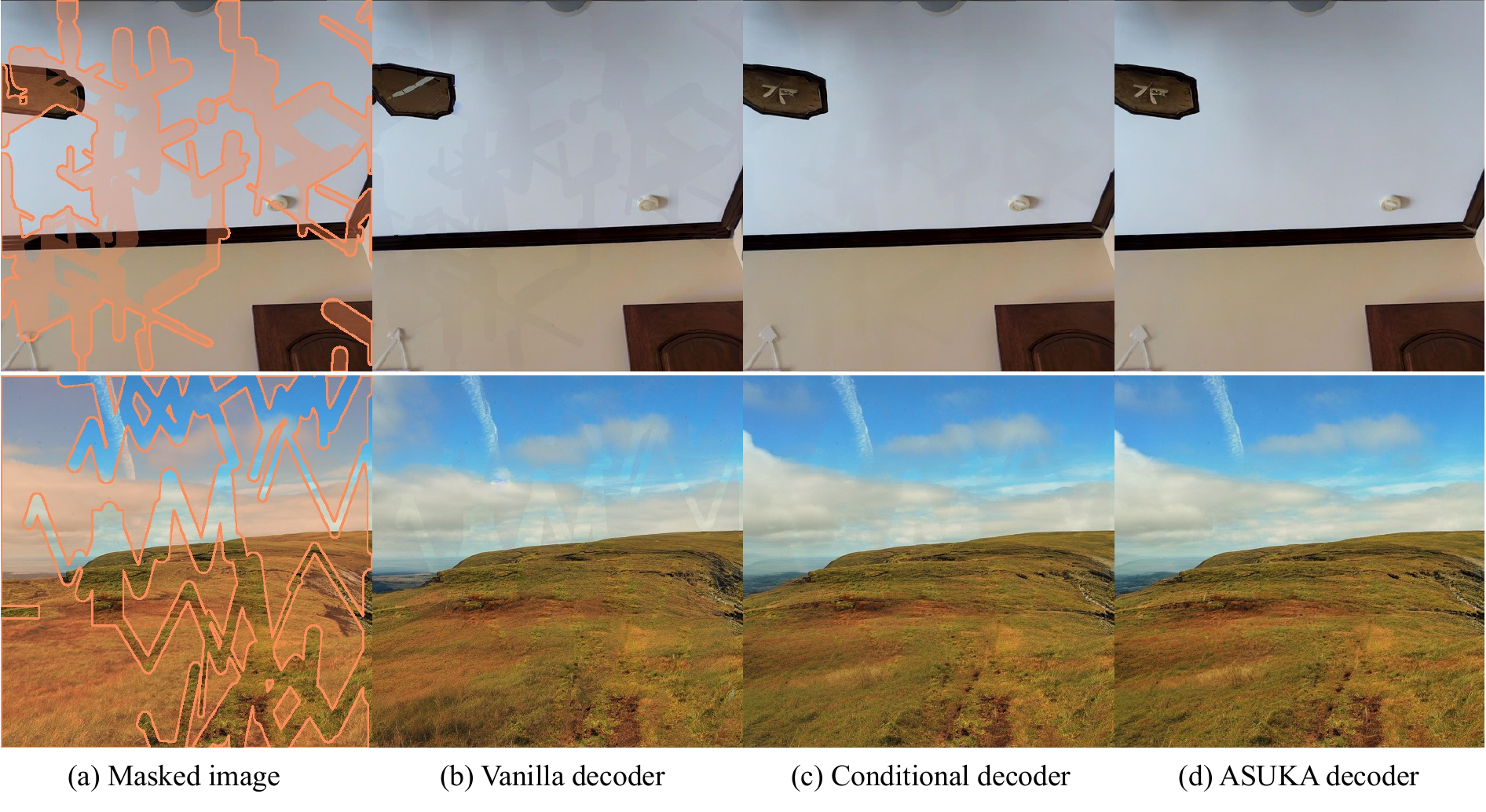}
\vspace{-0.3cm}
\caption{SD1.5 inpainting results decoded by (b) vanilla decoder of SD~\cite{Rombach_2022_CVPR}, (c) conditional decoder~\cite{zhu2023designing}, (d) our decoder.
Our decoder largely alleviate the mask-unmask color inconsistency.
\label{fig:decoder_compare}}
\vspace{-0.2cm}
\end{figure}

\subsubsection{Mask-Unmask Align during Decoding}
We propose to solve the color-inconsistency and ensure the mask-unmask alignment during VAE decoding.

\paragraph{Unmask-region conditioned decoder}
The basic solution is to incorporate the ground-truth unmasked region in the decoding, then we could have access to the unbiased color information.
Zhu \emph{et. al.}~\cite{zhu2023designing} adopts decoder with additional inputs of masked images.
However, it still fails to handle the incompatible color and texture between the original images and compressed ones in challenging scenes as verified in  Fig.~\ref{fig:decoder_compare} (c).
The gap between degraded and original images makes it challenging to explicitly address this issue.

\paragraph{Mask-unmask color-consistent decoder}
To train the decoder to ensure color-consistency between generated latent and unmasked pixels,
we re-formulate the decoding as a local harmonization task.
Our decoder involves additional inputs of masked images in the pixel-wise color space and the 0-1 mask.
To properly train the decoder,
we propose the color and latent augmentation as shown in  Fig.~\ref{fig:aug} to estimate and enlarge the color-inconsistency. 
We follow the standard VAE training pipeline, but replacing the inputs with augmented ones.
Particularly, we use the original image as the reconstruction target and use color and latent augmentation to corrupt input image, simulating the information loss of VAE and domain gap between generated and real latent, respectively.
This forces the decoder to reconstruct the clean image based on the ground-truth unmasked region.

\paragraph{Color augmentation} 
We use color augmentation to capture the VAE loss as in Fig.~\ref{fig:decoder_compare} (b).
Empirically, further conditioned on unmasked image alleviate but not solve the color inconsistency issue, as shown in Fig.~\ref{fig:decoder_compare} (c).
Hence, we need to explicitly train the decoder to ensure color consistency.
To this end, we augment all training images in brightness, contrast, saturation, and hue, and requires the decoder to reconstruct original image conditioned on the unaugmented unmasked image.
This encourages the decoder to faithfully follow the unmasked regions.

\paragraph{Latent augmentation} 
To simulate the gap between generated and real latent, we incorporate the artifacts generated from the generative models to train the decoder. 
However, denoising to real images iteratively is notably time-consuming, even with DDIM~\cite{song2021denoising}. 
To balance the efficiency and 
efficacy, we design a one-step estimation.
As our target is to capture the generation gap, we use the clean latent $\bm{z}_0$ and all-zero mask $\mathbf{M}$ as conditions.
This tells the generator all the needed information to generate the clean latent, ensuring the generated latent preserves content and only shift from the generation gap.
We follow the standard pipeline to estimate $\bm{z}_0$ with modified conditions as:
\begin{equation}
\hat{\bm{z}}_0=\frac{1}{a}(\bm{z}_t-b\varepsilon_\theta\left([\bm{z}_t;\bm{z}_0;\mathbf{M}],t\right)),
\label{eq:z0_pred}
\end{equation}
where the timestep $t$ is randomly sampled from $[500,1000)$; $a$ indicates the  prescribed variance schedule, $a^2+b^2=1$ in diffusion models while $a+b=1$ in rectified flow models; $\varepsilon_\theta(\cdot)$ is the frozen generator take as inputs noised $\bm{z}_t$, unmasked $\bm{z}_0$, and all-zero masking $\mathbf{M}$. 
The large step denoising is chosen to increase the distribution gap, as empirically the generator could produce reliable results in small $t$ given the unmasked latent condition $\bm{z}_0$.
Then we decode $\hat{\bm{z}}_0$ to image as the latent augmented inputs.
This makes latent augmentation an off-line strategy.
We apply latent augmentation to 50\% training images. 
The fine-tuned decoder showcases superior consistency as compared in Fig.~\ref{fig:decoder_compare}.

\begin{figure*}
\centering
\includegraphics[width=\linewidth]{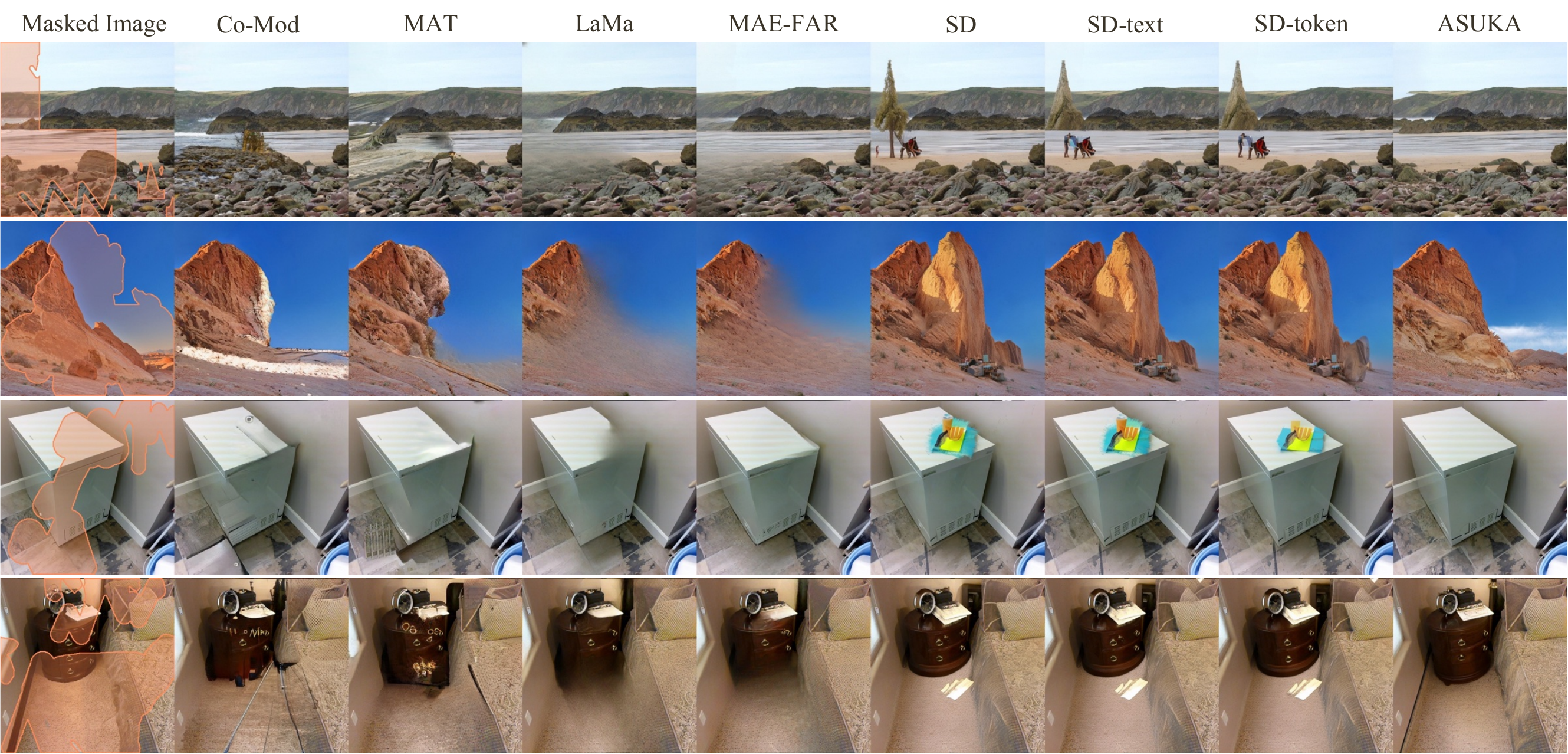}
\caption{
Inpainting results for 512$^2$ images.
GANs generate blurred results; SD variants hallucinate unreasonable objects and suffer from color shift.
ASUKA achieves unwanted-object-mitigated and color-consistent inpainting.
More results are in the appendix.
}
\label{fig:main-result}
\end{figure*}

\section{Experiments}

\paragraph{Evaluation datasets} 
We follow previous works to evaluate on the standard benchmark Places 2~\cite{zhou2017places} validation set of 36,500 images.
In addition, to validate across different domains and mask styles, we construct a evaluation dataset, dubbed as MISATO, from \underline{M}atterport3D~\cite{Matterport3D}, 
Fl\underline{i}ckr-Land\underline{s}cape~\cite{lin2021infinity},
Meg\underline{a}Dep\underline{t}h~\cite{MDLi18}, and C\underline{O}CO 2014~\cite{lin2014microsoft} to handle indoor, outdoor, building and background inpainting, respectively.
We select 500 representative examples of size $512^2$  and $1024^2$ from each dataset, forming a total of 2,000 testing examples.
See details in the appendix.

\paragraph{General evaluation metrics}
We use the
Learned Perceptual Image Patch Similarity (LPIPS)~\cite{zhang2018perceptual} to calculate the patch-level image distances,
Fr\'echet Inception Distance (FID)~\cite{heusel2017gans} to compare the distribution distance between generated images and real images, and Paired/Unpaired Inception Discriminative Score (P-IDS/U-IDS)~\cite{zhao2021large} to measure the human-inspired linear separability.

\paragraph{Evaluate object hallucination and color-consistency}
We introduce two new metrics to assess the  object hallucination and color-consistency of inpainted images.
(1) \emph{CLIP@mask} (C@m): We use CLIP to get visual features from both the ground-truth and the inpainted masked region, then calculate their cosine similarity. 
Following the standard CLIP score, we multiply the result by 100 and clip negative values, yielding a range from 0 to 100.
(2) \emph{Gradient@edge} (G@e): We calculate the average pixel gradient difference along the edges of the masked region with respect to the ground-truth image to assess color smoothness. A smaller gradient difference means more similar color transitions  to the ground-truth image and, therefore, less color shift.

\begin{table*}
\centering
\renewcommand\tabcolsep{2.0pt}
\begin{threeparttable}
\caption{Quantitative comparison on MISATO and Places 2.
Top-3 results are colored.
}
\small{
\begin{tabular*}{\textwidth}{@{\extracolsep{\fill}}lcccccccccccc}
\toprule
Dataset & \multicolumn{6}{c}{MISATO}  & \multicolumn{6}{c}{Places 2 }  \\
Method & LPIPS$\downarrow$ & FID$\downarrow$ & U-IDS$\uparrow$ & P-IDS$\uparrow$  & C@m$\uparrow$ & G@e$\downarrow$ & LPIPS$\downarrow$ & FID$\downarrow$ & U-IDS$\uparrow$ & P-IDS$\uparrow$   & C@m$\uparrow$ & G@e$\downarrow$\\
\midrule
\midrule
Co-Mod~\cite{zhao2021large} & 0.179& 17.421& 0.243& 0.109 & 0.924 & 52.106 & 0.267 &  5.794 & 0.274 & 0.096 & 0.951 & 166.914\\
MAT~\cite{li2022mat} & 0.176 & 17.261 & 0.255 & 0.122 & 0.925 & 48.722 & 0.202 & 3.765 & 0.348 & 0.195 & 0.955 & 163.442\\
LaMa~\cite{suvorov2022resolution} & \cellcolor{rank3}0.155 & 15.436 & 0.260 & 0.135 & 0.928 & \cellcolor{rank2}46.270 & 0.202 & 6.693 & 0.247 & 0.050 & 0.953 & 153.653\\
MAE-FAR~\cite{cao2022learning}  & 0.142\cellcolor{rank1} & 13.283 & 0.282 & 0.153 & 0.940 & \cellcolor{rank1}43.613 & \cellcolor{rank1}0.174 & 3.559 & 0.307 & 0.105 & 0.958 & 149.843\\
SD-Repaint~\cite{lugmayr2022repaint} & 0.227 & 27.861 & 0.016 & 0.007 & 0.915 & 80.410 & 0.251 & 12.466 & 0.217 & 0.045 & 0.947 & 176.421\\
SD~\cite{Rombach_2022_CVPR} & 0.168 & 12.812 & 0.345 & 0.211 & 0.951 & 63.844 & 0.193 & 1.514 & 0.375 & \cellcolor{rank3}0.207 & \cellcolor{rank3}0.959 & 160.705\\
SD-text & 0.164 & 12.603 & 0.337 & \cellcolor{rank3}0.207 & 0.952 & 63.776 & 0.191 & \cellcolor{rank3}1.506 & 0.373 & 0.202 & \cellcolor{rank3}0.959 & 160.418\\
\midrule
SD-token~\cite{wang2024repositioning} & 0.160 & 12.517 & 0.331 & 0.204 & \cellcolor{rank3}0.955 & 61.700 & 0.189 & \cellcolor{rank2}1.477 & \cellcolor{rank2}0.390 & \cellcolor{rank2}0.234 & \cellcolor{rank2}0.960 & 158.924\\
SD-IP~\cite{ye2023ip-adapter} & 0.157 & \cellcolor{rank3}12.204 & \cellcolor{rank2}0.398 & \cellcolor{rank3}0.242 & \cellcolor{rank2}0.956 & 62.704 & \cellcolor{rank3}0.186 & 1.539 & \cellcolor{rank3}0.389 & 0.173 & 0.953 & 148.571\\
SD-T2I~\cite{mou2024t2i} & 0.166 & 13.806 & 0.365 & 0.222 & 0.949 & 63.866 & 0.195 & 1.720 & 0.384 & 0.160 & 0.951 & \cellcolor{rank3}148.549\\
SD-CAEv2~\cite{zhang2023cae} & 0.157 & 29.179 & 0.193 & 0.045 & 0.901 & 69.890 & 0.192 & 6.887 & 0.287 & 0.065 & 0.921 & 151.863 \\
SD-LaMa~\cite{suvorov2022resolution} & 0.157 & \cellcolor{rank2}12.159 & \cellcolor{rank3}0.390 & \cellcolor{rank2}0.256 & \cellcolor{rank2}0.956 & 62.726 & 0.188& 1.522 & 0.389 & 0.168 & 0.953 & \cellcolor{rank2}148.461\\
\midrule
ASUKA-SD & \cellcolor{rank2}0.150 & 
\cellcolor{rank1}11.495 & \cellcolor{rank1}0.423 & \cellcolor{rank1}0.312 & \cellcolor{rank1}0.958 & \cellcolor{rank3}47.753 & \cellcolor{rank2}0.183 & \cellcolor{rank1}1.230 & \cellcolor{rank1}0.413 & \cellcolor{rank1}0.287 & \cellcolor{rank1}0.961 & \cellcolor{rank1}147.733\\
\bottomrule
\end{tabular*}
}
\label{tab:quantitative}
\end{threeparttable}
\vspace{-0.5cm}
\end{table*}

\paragraph{Competitors}
We primarily use the SD v1.5 inpainting model~\cite{Rombach_2022_CVPR} to analyze and compare ASUKA with competitors, while validating ASUKA's generalization ability with FLUX.
We consider three SD v1.5 inpainting variants:
SD: uses a null-prompt for unconditional generation;
SD-text: uses "background" as a prompt since no captions are used in inpainting;
SD-token~\cite{wang2024repositioning}: uses learnable tokens trained with ASUKA's pipeline.
To test other ways of incorporating the MAE condition, we implement the following:
SD-IP, uses IP-Adapter~\cite{ye2023ip-adapter};
SD-T2I, uses T2I-Adapter~\cite{mou2024t2i};
SD-CAEv2, uses a CLIP-style alignment module CAEv2~\cite{zhang2023cae};
We also test SD-LaMa, which inputs LaMa~\cite{suvorov2022resolution} inpainting results instead of MAE.
We also compare with leading inpainting algorithms Co-Mod~\cite{zhao2021large}, MAT~\cite{li2022mat}, LaMa~\cite{suvorov2022resolution}, MAE-FAR~\cite{cao2022learning}, and SD-Repaint~\cite{lugmayr2022repaint}.
We provide implementation details in the appendix.

\subsection{Comparison on Benchmarks}

\paragraph{Quantitative comparison} Results on SD are reported in Tab.~\ref{tab:quantitative}.
\textbf{(1)}: Although ASUKA-SD is based on a fixed SD model, it consistently outperforms SD across all evaluation metrics, achieving state-of-the-art results in FID, U-IDS, and P-IDS. Notably, U-IDS and P-IDS are closely aligned with human preferences~\cite{zhao2021large} and have a potential maximum score of 0.5, highlighting ASUKA's strong performance.
\textbf{(2)}: Compared to other adapters that align the MAE prior with SD, ASUKA-SD shows consistently superior performance across all metrics. This demonstrates the effectiveness of our straightforward alignment module.
\textbf{(3)}: While the LaMa condition improves inpainting quality, as shown by FID and IDS scores, it is less effective than the MAE condition. When using the MAE condition as a prior, improvements can be attributed to better mitigation of object hallucination.
\textbf{(4)}: ASUKA-SD consistently performs better than all competitors on CLIP@mask, showcasing the strength of its improved mitigation of object hallucination.
\textbf{(5)}: Pixel-based GAN inpainting models perform better in the Gradient@edge metric, suggesting that color shifts may originate from the compressed latent space. ASUKA-SD, however, still shows significant improvements over all SD variants, highlighting its enhanced color consistency.
\textbf{(6)}: The second-to-best LPIPS scores are partially due to using a frozen SD, where ASUKA achieves consistent improvements but remains constrained by the frozen U-Net.
These results confirm that ASUKA-SD improves color consistency and mitigation of object hallucination in inpainting, even when using frozen latent inpainting models. This advantage is evident both in the in-distribution dataset Places2 and the out-of-distribution dataset MISATO.

\paragraph{Qualitative comparison} examples are shown in Fig.~\ref{fig:main-result}.
\textbf{(1)} The state-of-the-art inpainting algorithms usually suffer from unnatural generation, for example the unnatural boundaries in the third and fourth rows, and failed inpainting of tower in the third-to-last row.
LaMa and MAE-FAR sometimes lead to blurred inpainting results, especially in the scenario of large continuous masks.
\textbf{(2)} The SD variants usually suffer from the unwanted object insertion issue and hallucinate unreasonable objects, in almost all the illustrated images.
\textbf{(3)} In contrast, ASUKA enjoys unwanted-object-mitigated and color-consistent inpainting.

\subsection{Further Analysis of ASUKA}
In this part, we conduct more experiments to analysis ASUKA.
More analysis can be found in the appendix.

\paragraph{Extension to FLUX}
To demonstrate ASUKA's versatility, we trained it on FLUX (see Tab.~\ref{tab:flux}). 
ASUKA-FLUX consistently outperforms the original FLUX. 
Results on Places 2 and qualitative comparisons are in the appendix.

\begin{table}
\centering
\small
\begin{threeparttable}
\caption{\label{tab:flux} FLUX and ASUKA-FLUX on MISATO@512. Results on 1K and qualtitative results are in the appendix.
}
\vspace{-0.2cm}
\renewcommand\tabcolsep{1.5pt}
\begin{tabular*}{\linewidth}{@{\extracolsep{\fill}}lcccccc}
\toprule
Decoder & LPIPS$\downarrow$  & FID$\downarrow$ & U-IDS$\uparrow$ & P-IDS$\uparrow$ & C@m$\uparrow$ & G@e$\downarrow$
\\
\midrule
FLUX & 0.254 & 12.839 & 0.351 & 0.223 & 0.951 & 65.928 \\
ASUKA-FLUX & \textbf{0.206} & \textbf{11.372} & \textbf{0.428} & \textbf{0.327} & \textbf{0.962} & \textbf{48.635}\\
\bottomrule
\end{tabular*}
\end{threeparttable}
\vspace{-0.2cm}
\end{table}

\begin{table}
\centering
\small
\begin{threeparttable}
\caption{\label{tab:decoder-ablation}Comparison of different decoders for SD.
}
\renewcommand\tabcolsep{2.0pt}
\begin{tabular*}{\linewidth}{@{\extracolsep{\fill}}lcccccc}
\toprule
Decoder & LPIPS$\downarrow$  & FID$\downarrow$ & U-IDS$\uparrow$ & P-IDS$\uparrow$ & C@m$\uparrow$ & G@e$\downarrow$
\\
\midrule
VAE  & 0.156 & 11.949 & 0.387 & 0.253 & 0.953 & 63.142\\
+ cond.  & 0.151 & 11.634 & 0.410 & 0.272 & 0.955 & 48.588\\
+ color  & 0.152&11.603 & 0.407 & 0.273 & 0.954 & 49.538\\
\midrule
Ours& \textbf{0.150} & \textbf{11.495} & \textbf{0.423} & \textbf{0.312} & \textbf{0.958} & \textbf{47.753}\\\bottomrule
\end{tabular*}
\end{threeparttable}
\end{table}

\paragraph{Ablation of decoder}
For the decoder, we compare ASUKA-SD with 
(1) \textit{VAE}: the decoder used in SD;
(2) \textit{+ cond.}: the decoder conditioned on unmasked image~\cite{zhu2023designing};
(3) \textit{+ color}: only trained with color augmentation ;
Results are in Tab.~\ref{tab:decoder-ablation}, showing the superiority of our decoder.

\begin{table}
\centering
\begin{threeparttable}
\renewcommand\tabcolsep{2.0pt}
\small
\caption{\label{tab:align-ablation}Ablation of different alignment modules.
}
\begin{tabular*}{\linewidth}{@{\extracolsep{\fill}}lcccccc}
\toprule
Align& LPIPS$\downarrow$  & FID$\downarrow$ & U-IDS$\uparrow$ & P-IDS$\uparrow$& C@m$\uparrow$ & G@e$\downarrow$\\
\midrule
linear & 0.155 & 11.934 & 0.400 & 0.263 & 0.953 & 48.983\\
attn  & 0.152 & 11.613 & 0.403 & 0.268 & 0.954 & 48.785\\
cross x4 & 0.152 & 11.762 & 0.405 & 0.256 & 0.953 & 48.279 \\
\midrule
Ours &\textbf{0.150} & \textbf{11.495} & \textbf{0.423} & \textbf{0.312} & \textbf{0.958} & \textbf{47.753}\\
\bottomrule
\end{tabular*}
\end{threeparttable}
\vspace{-0.2cm}
\end{table}

\paragraph{Ablation of alignment module}
We validate the efficacy of our alignment module step by step:
(1) \textit{linear}: Use linear layer to align feature dimension only;
(2) \textit{attn}: Based on \textit{linear}, further use a single self-attention block to align the distribution;
(3) \textit{cross x4}: we instead use learnable query and 4 cross-attention layers to learn the MAE prior.
ASUKA-SD adopts 4 self-attention blocks.
Results are shown in Tab.~\ref{tab:align-ablation}. 
The self-attention block shows improved results compared with only align dimension and cross-attention block.
Using 4 self-attention blocks improves the capacity.

\section{Conclusion}
In this paper, we proposed Aligned Stable inpainting with Unknown Areas prior (ASUKA) to achieve unwanted-object-mitigated and color-consistent inpainting via frozen latent inpainting models.
To avoid unwanted object insertion, we adopt a reconstruction-based masked auto-encoder (MAE) as the context-stable prior for masked region purely from unmasked region.
Then we align the context-stable prior to frozen generative models with the proposed alignment module.
To achieve color-consistency, we resolve the mask-unmask color inconsistency in the latent decoding process.
We train an unmask-region conditioned VAE decoder to perform local harmonization during the decoding process.
To validate the efficacy of inpainting algorithms in different image domains and mask types, we introduce an evaluation dataset, named as MISATO, from existing datasets.
We propose two new metrics to explicitly evaluate the object hallucination and color-consistency of inpainted images.
ASUKA enjoys unwanted-object-mitigated and color-consistent inpainting results and superior than leading inpainting models.

\paragraph{Acknowledgments}
The authors would like to thank Huawei Ascend Cloud Ecological Development Project for the support of Ascend 910 processors.

{
    \small
    \bibliographystyle{ieeenat_fullname}
    \bibliography{main}
}

\clearpage
\setcounter{page}{1}
\maketitlesupplementary

\section{Brief Introduction of Backbone Models}
We evaluate our proposed solution on two inpainting models: the Stable Diffusion v1.5 inpainting model (SD)~\cite{Rombach_2022_CVPR} and the Control-Net fine-tuned FLUX inpainting model (FLUX)~\cite{flux-inpainting}. 
Both models are representative latent inpainting models that use a VAE~\cite{kingma2013auto} to compress images into a smaller latent space.  
In SD, a diffusion process~\cite{sohl2015deep} maps the latent space to random Gaussian noise, and a U-Net~\cite{ronneberger2015u} learns the reverse denoising path. 
Text condition is introduced through cross-attention layers~\cite{Vaswani2017Attention}. 
The inpainting version of SD extends the U-Net input by concatenating the masked image and mask with the noise along the channel dimension.
Conversely, FLUX uses rectified flow~\cite{liu2022flow,albergo2022building,lipman2022flow} to map the latent space to noise and a vision transformer~\cite{peebles2023scalable} for generation. 
Text condition is applied by concatenating text with image patches as transformer input, while a pooled text condition is injected into the normalization layers. 
Since the original FLUX~\cite{flux} does not support inpainting, we use a Control-Net~\cite{zhang2023adding} fine-tuned version~\cite{flux-inpainting} that modifies FLUX’s transformer output by conditioning on the masked image and mask.
We demonstrate that our ASUKA effectively improves unwanted object mitigation and color consistency of these  models.

\section{Details about MISATO}
The principle of constructing MISATO is to select the most representative and diverse examples.
To this end, for first three datasets, we use CLIP visual model~\cite{radford2021learning} to extract semantic visual features. 
Then we use BisectingKMeans~\cite{steinbach2000comparison} to cluster each dataset into 500 clusters, and select the cluster centers as the evaluation data.
The selected data are center cropped and then resized to $512^2$.
For COCO, we focus on the background inpainting. 
To this end, for each data we identify the foreground with provided segmentation and remove it from the generated masks, yielding a dataset specified for purely background inpainting.

\begin{figure}
\includegraphics[width=\linewidth]{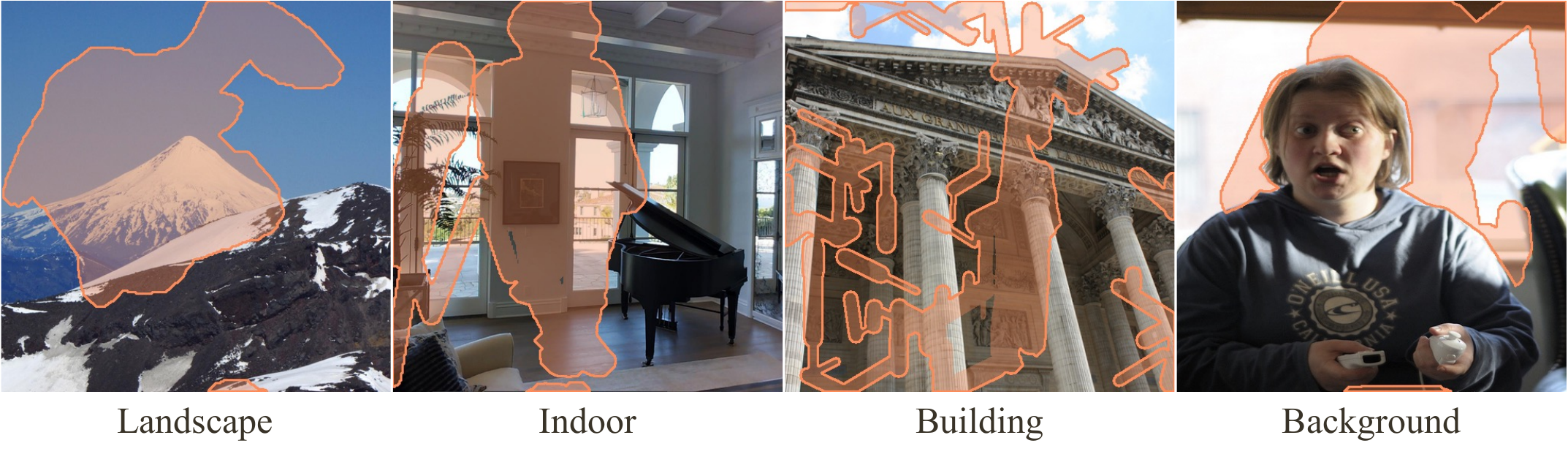}
\caption{Different image domains in MISATO.
\label{fig:misato}
}
\end{figure}

Combined together, MISATO contains 2000 examples from four inpainting domains, indoor, outdoor landscape, building, and background, as shown in Fig.~\ref{fig:misato}.
 we adopt the masking strategy as in Sec.~\ref{sec:mask-strategy} but excluding the rectangle and complement rectangle masks.
The masking ratio is set as $[0.2, 0.8]$.

\begin{figure*}
\centering
\includegraphics[width=\linewidth]{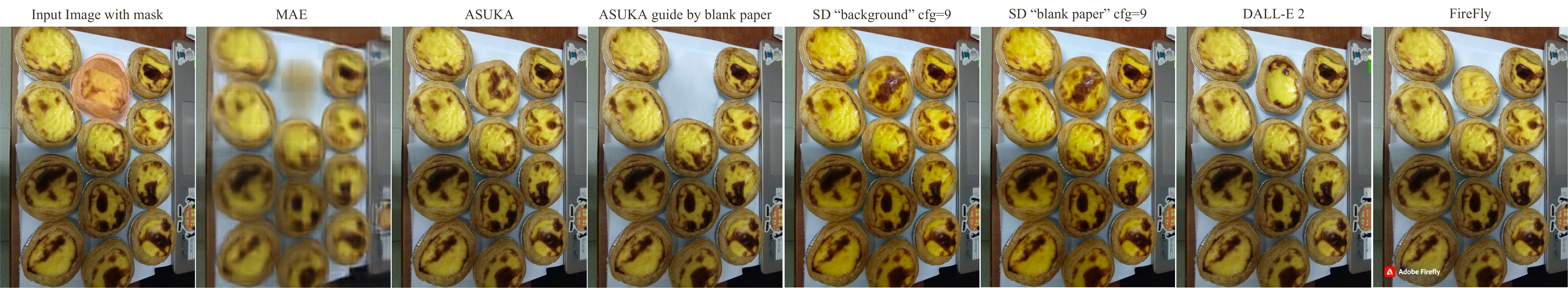}
\caption{
The curse of self-attention, causing the MAE falsely estimate the masked region and powerful text-guided diffusion models fail to generation content based on text prompts.
ASUKA potential circumvents this issue by using a blank paper image as the input to the MAE to provide correct prior.
}
\label{fig:limit}
\end{figure*}

\section{Implementation Details}
We use Places2~\cite{zhou2017places} to train ASUKA. 
For the MAE~\cite{he2022masked} used in ASUKA, 
we train on images of size $256^2$, which is efficient and produce context-stable guidance for generative models to generate high-resolution images.
We fine-tune the MAE with a batch size of 1024.
We train the alignment module with AdamW~\cite{loshchilov2018decoupled} of learning rate 5e-2 with the standard diffusion objective.
We set $p$ as $100\%$ and linearly decay it to $10\%$ in the first 2K training steps and then freeze.
For SD's decoder, we fine-tune from~\cite{zhu2023designing} for 50K steps with a batch size of 40 and learning rate of 8e-5 with cosine decay. 
For FLUX's decoder, we fine-tune from the original decoder with the same setup.
We use ColorJitter for color augmentation, with brightness 0.15, contrast 0.2, saturation 0.1, and hue 0.03. 

\section{Further Analysis}

\begin{table}
\centering
\renewcommand\tabcolsep{1.5pt}
\caption{Comparison of ASUKA with text-guided SD \label{tab:asuka-vs-text}}
\small
\begin{tabular*}{\linewidth}{@{\extracolsep{\fill}}lcccccc}
\toprule
Model  & LPIPS$\downarrow$  & FID$\downarrow$ & U-IDS$\uparrow$ & P-IDS$\uparrow$& C@m$\uparrow$ & G@e$\downarrow$\\
\midrule
\midrule
SD (BLIP2) & 0.163  & 12.536 & 0.370 & 0.225 & 0.880 & 70.846 \\
ASUKA-SD &\textbf{0.150} & \textbf{11.495} & \textbf{0.423} & \textbf{0.312} & \textbf{0.958} & \textbf{47.753}\\
\bottomrule
\end{tabular*}
\end{table}
\paragraph{Comparison with text-guided inpainting}
We compare ASUKA with text-guided SD model, as shown in Tab.~\ref{tab:asuka-vs-text}.
We run SD inpainting sing text captions generated by BLIP2~\cite{li2023blip}.
ASUKA performs better, since captions describe the entire image, while MAE focuses on reconstructing only the masked region, leading to more precise guidance.

\begin{table}
\centering
\caption{Ablation of $p$ \label{tab:ablate-p}}
\small
\renewcommand\tabcolsep{1.5pt}
\begin{tabular*}{\linewidth}{@{\extracolsep{\fill}}lcccccc}
\toprule
Model  & LPIPS$\downarrow$  & FID$\downarrow$ & U-IDS$\uparrow$ & P-IDS$\uparrow$& C@m$\uparrow$ & G@e$\downarrow$\\
\midrule
\midrule
p=0 & 0.155 & 11.804 & 0.403 & 0.288 & 0.940 & 48.032 \\
p=1 & 0.152 & 11.734 & 0.394 & 0.296 & 0.947 & 47.997\\
linear decay p & 0.152 & 11.558 & 0.405 & 0.307 & 0.955 & 47.814\\
Ours &\textbf{0.150} & \textbf{11.495} & \textbf{0.423} & \textbf{0.312} & \textbf{0.958} & \textbf{47.753}\\
\bottomrule
\end{tabular*}
\end{table}
\paragraph{Ablation of $p$}
We analyze how different values of $p$ affect ASUKA in Tab.~\ref{tab:ablate-p}. 
The results show that our warm-up and freeze strategy outperforms other approaches.

\begin{table}
\centering
\caption{Additional results on benchmark datasets \label{tab:celeba-ffhq}}
\footnotesize
\renewcommand\tabcolsep{1.5pt}
\begin{tabular*}{\linewidth}{@{\extracolsep{\fill}}llcccccc}
\toprule
Dataset & Model  & LPIPS$\downarrow$  & FID$\downarrow$ & U-IDS$\uparrow$ & P-IDS$\uparrow$& C@m$\uparrow$ & G@e$\downarrow$\\
\midrule
\midrule
\multirow{2}{*}{CelebA-HQ} &
SD & 0.132 & 11.968 & 0.282 & 0.101 & 0.939 & 42.870\\
&ASUKA-SD & \textbf{0.129} & \textbf{10.190} & \textbf{0.293} & \textbf{0.134} & \textbf{0.941} & \textbf{40.503} \\
\bottomrule
\multirow{2}{*}{FFHQ} &
SD & 0.139 & 2.235 & 0.371 & 0.197 & 0.944 & 43.529\\
&ASUKA-SD & \textbf{0.131} & \textbf{2.060} & \textbf{0.386} & \textbf{0.205} & \textbf{0.955} & \textbf{30.848}  \\
\bottomrule
\end{tabular*}
\end{table}
\paragraph{Additional Results}
We further compare ASUKA with standard SD on two additional datasets: CelebA-HQ~\cite{karras2018progressive} and FFHQ~\cite{karras2019style}. As shown in Tab.~\ref{tab:celeba-ffhq}, these results provide more evidence of ASUKA’s effectiveness.

\begin{table}
\centering
\caption{Our Decoder in Text-Guided Inpainting.\label{tab:decoder-t2i}}
\small
\renewcommand\tabcolsep{1.5pt}
\begin{tabular*}{\linewidth}{@{\extracolsep{\fill}}lcccccc}
\toprule
Model  & CLIPScore$\uparrow$ &  LPIPS$\downarrow$  & FID$\downarrow$ & U-IDS$\uparrow$& C@m$\uparrow$ & G@e$\downarrow$\\
\midrule
\midrule
SD & 0.297 & 0.180 & 30.255 & 0.312 & 0.930 & 57.136 \\
ASUKA-SD & \textbf{0.298} & \textbf{0.175} & \textbf{29.350} & \textbf{0.350} & \textbf{0.931} & \textbf{38.123} \\
\bottomrule
\end{tabular*}
\end{table}
\paragraph{Our Decoder in Text-Guided Inpainting}
To test the generalizability of our decoder, we evaluate it on text-guided inpainting tasks. 
We compare our decoder with the original SD decoder using 1,000 randomly sampled images from “jackyhate/text-to-image-2M”~\cite{zk_2024}. 
The results in Tab.~\ref{tab:decoder-t2i} confirm its effectiveness for general inpainting tasks.

\begin{table}
\centering
\caption{Effect of each module.\label{tab:effect-module}}
\small
\renewcommand\tabcolsep{1.5pt}
\begin{tabular*}{\linewidth}{@{\extracolsep{\fill}}lcccccc}
\toprule
MAE  & LPIPS$\downarrow$  & FID$\downarrow$ & U-IDS$\uparrow$ & P-IDS$\uparrow$& C@m$\uparrow$ & G@e$\downarrow$\\
\midrule
\midrule
SD w/ MAE & 0.157 & 12.093 & 0.397 & 0.236 & 0.953 & 62.845  \\
SD w/ decoder & 0.159 & 12.075 & 0.411 & 0.283 & 0.954 & 49.376 \\
ASUKA-SD &\textbf{0.150} & \textbf{11.495} & \textbf{0.423} & \textbf{0.312} & \textbf{0.958} & \textbf{47.753}\\
\bottomrule
\end{tabular*}
\end{table}
\paragraph{Ablation on independent modules}
To understand the contribution of each module in ASUKA, we evaluate SD with the proposed modules added separately. The results, shown in Tab.~\ref{tab:effect-module}, highlight the effectiveness of each module.

\begin{table}
\centering
\caption{Comparison of ASUKA using pre-trained MAE v.s. fine-tuned MAE.\label{tab:pt-vs-ft}}
\begin{tabular*}{\linewidth}{@{\extracolsep{\fill}}lcccc}
\toprule
MAE  & LPIPS$\downarrow$  & FID$\downarrow$ & U-IDS$\uparrow$ & P-IDS$\uparrow$\\
\midrule
\midrule
pre-trained & 0.151 & 11.513 & 0.354 & \textbf{0.258}\\
fine-tuned  & \textbf{0.150} & \textbf{11.460} & \textbf{0.368} & 0.256\\
\bottomrule
\end{tabular*}
\end{table}

\paragraph{Ablation of MAE prior}
We compare our fine-tuned MAE with directly adopting the MAE trained in~\cite{cao2022learning}.
To this end, we train ASUKA with the MAE in~\cite{cao2022learning} using the same training strategy and compare the results in Tab.~\ref{tab:pt-vs-ft}.
Results suggest the improvements of fine-tuning MAE, especially on FID and U-IDS.
This improvement comes from the better adaptation on the real-world masks.
\begin{table}
\centering
\begin{threeparttable}
\caption{\label{tab:user-study}User-study of top-1 ratio among all the inpainting results. 
}
\begin{tabular*}{\linewidth}{@{\extracolsep{\fill}}lcc}
\toprule
Model & UOM (\%) & CC(\%) 
\\
\midrule
\midrule
Co-Mod~\cite{zhao2021large} & 3.98  & 4.98  \\
MAT~\cite{li2022mat} & 7.40 & 3.20\\
LaMa~\cite{suvorov2022resolution} & 8.18 & 8.28\\
MAE-FAR~\cite{cao2022learning} & 4.88 & 5.60\\
\midrule
SD~\cite{Rombach_2022_CVPR} & 10.58 & 5.75\\
SD-text  & 7.70& 15.83 \\
SD-prompt  & 16.18 & 15.78 \\
SD-Repaint~\cite{lugmayr2022repaint} & 1.60 & 0.55\\
\midrule
ASUKA-SD & \textbf{39.43} & \textbf{40.05} \\
\bottomrule
\end{tabular*}
\end{threeparttable}
\end{table}%

\paragraph{User-study}
To evaluate the user preference on inpainting algorithms, we conduct an user-study.
Specifically, we randomly select 40 testing images.
We ask the user to select the best inpainting results from the following perspectives respectively:
i) Unwanted-object-mitigation (UOM): the generated region should be context-stable with surrounding unmasked region, with a preference of not generating new elements;
ii) Color-consistency (CC) : the color consistency between masked and unmasked regions.
We collect 100 valid anonymous questionnaire results, and report the average selection ratio among all the inpainting algorithms in Tab.~\ref{tab:user-study}.
This result validate the efficacy of ASUKA on alignment with human preference.

\paragraph{Limitation: The "curse" of self-attention}
The primary limitation of ASUKA stems from the inefficacy of the MAE prior, mainly due to issues within the self-attention module. 
Specifically, as shown in Fig.~\ref{fig:limit}, the presence of multiple similar objects in an image may lead the MAE to incorrectly predict a similar object in the masked region, conflicting with the goal of object removal. 
Notably, this curse of self-attention significantly impacts diffusion-based generative models. 
It results in the inability to accurately follow "blank paper" text prompts, even when employing a substantial classifier-free guidance scale of 9. 
This issue is not unique to SD but is also a common problem in other advanced text-guided diffusion models, such as OpenAI's DALL-E 2~\cite{ramesh2022hierarchical} and Adobe's FileFly~\cite{firefly}. 
Nevertheless, ASUKA has the potential to circumvent this issue by modifying the MAE prior, for instance, by instead using a blank paper image as the input to MAE prior. 
A more comprehensive solution would involve extra control on self-attention layers in diffusion models, which we leave as future work.

\paragraph{Potential negative impact}
As an image editing tool, our proposed ASUKA will generate images based on user intentions for masking specific parts of the image, potentially resulting in unrealistic renderings and posing a risk of misuse.

\end{document}